\documentclass[10pt,twocolumn,letterpaper]{article}

\usepackage[pagenumbers]{cvpr} 

\usepackage[dvipsnames]{xcolor}

\definecolor{cvprblue}{rgb}{0.21,0.49,0.74}
\usepackage[pagebackref,breaklinks,colorlinks,citecolor=cvprblue]{hyperref}

\usepackage{ulem}
\usepackage{multirow}
\usepackage{booktabs}
\usepackage{amsmath}
\usepackage{amssymb}
\usepackage{listings}
\usepackage{algpseudocode}

\def\paperID{4289} 
\def\confName{CVPR}
\def\confYear{2024}

\title{MindBridge: A Cross-Subject Brain Decoding Framework}

\author{Shizun Wang \quad Songhua Liu \quad Zhenxiong Tan \quad Xinchao Wang$^{\dagger}$ \\
National University of Singapore \\
{\tt\small \{shizun.wang, songhua.liu, zhenxiong\}@u.nus.edu, xinchao@nus.edu.sg}
}

\begin{document}

\twocolumn[{
\renewcommand\twocolumn[1][]{#1}%
\maketitle
\begin{center}
    \vspace{-5mm}
    \captionsetup{type=figure}
    \includegraphics[width=\textwidth]{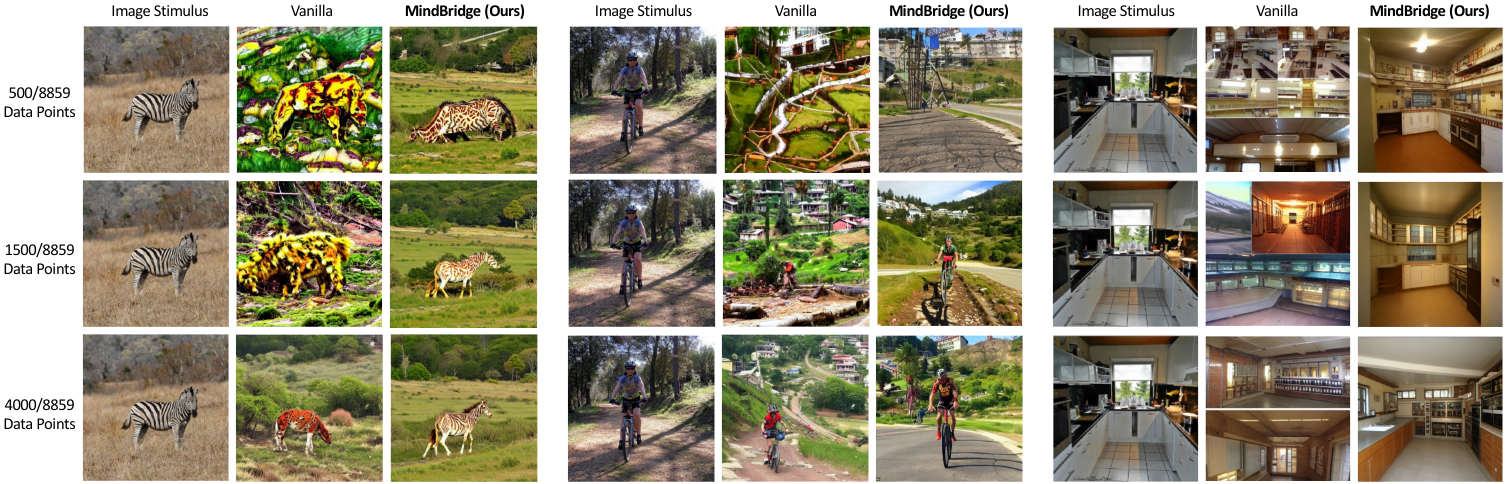}
    \captionof{figure}{\textbf{Image stimuli and images reconstructed from captured brain signals.} Given the limited training data from a new subject (subj07 from the NSD dataset \cite{allen2022massive}), our proposed \textbf{MindBridge} can faithfully reconstruct natural images using less data, benefiting from pretrained \textbf{cross-subject} knowledge. In contrast, the \textbf{Vanilla} method, which represents current methods following a \textbf{per-subject-per-model} paradigm, fails to learn effectively from limited data.}
\label{fig:teaser}
\end{center}
}]

\renewcommand{\thefootnote}{\fnsymbol{footnote}}
\footnotetext[2]{Corresponding author.}
\renewcommand{\thefootnote}{\arabic{footnote}}

\begin{abstract}
Brain decoding, a pivotal field in neuroscience, aims to reconstruct stimuli from acquired brain signals, primarily utilizing functional magnetic resonance imaging (fMRI). Currently, brain decoding is confined to a 
\textbf{per-subject-per-model} paradigm, limiting its applicability to the same individual for whom the decoding model is trained. This constraint stems from three key challenges: 1) the inherent variability in input dimensions across subjects due to differences in brain size; 2) the unique intrinsic neural patterns, influencing how different individuals perceive and process sensory information; 3) limited data availability for new subjects in real-world scenarios hampers the performance of decoding models.

In this paper, we present a novel approach, \textbf{MindBridge}, that achieves \textbf{cross-subject brain decoding} by employing only one model. Our proposed framework establishes a generic paradigm capable of addressing these challenges by introducing biological-inspired aggregation function and novel cyclic fMRI reconstruction mechanism for subject-invariant representation learning. 
Notably, by cycle reconstruction of fMRI, MindBridge can enable novel fMRI synthesis, which also can serve as pseudo data augmentation.
Within the framework, we also devise a novel reset-tuning method for adapting a pretrained model to a new subject. Experimental results demonstrate MindBridge's ability to reconstruct images for multiple subjects, which is competitive with dedicated subject-specific models. Furthermore, with limited data for a new subject, we achieve a high level of decoding accuracy, surpassing that of subject-specific models.
This advancement in cross-subject brain decoding suggests promising directions for wider applications in neuroscience and indicates potential for more efficient utilization of limited fMRI data in real-world scenarios.
Project page: \url{https://littlepure2333.github.io/MindBridge}
\end{abstract}

\vspace{-4mm}
\section{Introduction}
\label{sec:intro}

The human brain, an intricate web of neurons, possesses the remarkable ability to encode the sensory stimuli that we encounter every day, making sense of our perceptual world. While the reverse process, known as brain decoding, aims to reconstructs image stimulus from brain signals, which are primarily captured by functional magnetic resonance imaging (fMRI).
Brain decoding has been a subject of intense interest, as it offers the tantalizing prospect of unraveling the secrets of cognition and perception, and present a potential advancement for brain-computer interface (BCI) \cite{du2022fmri} and beyond.
From GANs \cite{shen2019deep, ozcelik2022reconstruction} to diffusion models \cite{takagi2022high, ozcelik2022reconstruction, scotti2023reconstructing}, the use of increasingly powerful generative models has enabled brain decoding to reconstruct more realistic and faithful images.
However, nowadays brain decoding is confronted with significant challenges that hinder its application on a broader scale.

Specifically, the current practice of brain decoding is confined to subject-specific applications \cite{takagi2022high, scotti2023reconstructing, gu2022decoding, mai2023unibrain, ozcelik2023brain, xia2023dream}. In other words, a decoding model trained on a one subject's brain can only be effectively applied to that same subject, and can not be applied to other subjects, which results in high expense of model storage and training. This limitation motivates us to move towards \textbf{cross-subject brain decoding}, which is capable of using one model to decode brain signals from multiple subjects and adapting to new subjects. Such a paradigm thereby can expand the utility of brain decoding in a more general way and bring more applicability in real scenarios.

However, it requires adequately addressing various substantial challenges in pursuit of this beautiful vision:
\textbf{1) Size Variability}: fMRI signals exhibit substantial size differences across subjects
, largely due to the inherent variability in brain size and structure, which necessitates a flexible approach to handle this variability.
\textbf{2) Diverse Neural Responses}: The intricacies of the brain extend beyond structural variations.
The way each subject's brain processes stimuli is uniquely shaped by their experiences, biases, and cognitive patterns, posing a challenge in unifying the interpretation of brain signals.
\textbf{3) Data Scarcity for New Subjects}: 
In real-world applications, it is highly cumbersome to acquire extensive fMRI data for new subjects if not infeasible at all. The high costs in both resources and time significantly hinder the training and adaptation of brain-decoding models for new subjects.

To address these challenges, we devise \textbf{``MindBridge"}, a novel framework designed to achieve cross-subject brain decoding. MindBridge employs innovative strategies to tackle each of the identified obstacles:
\textbf{1) Adaptive Signal Aggregation}: Inspired by neural-science findings that the brain activation is sparse and only neurons exceeding a certain threshold activate, we propose to use an aggregation function based on adaptive max pooling to extract most useful information
, and unify the input dimension of fMRI signals across different subjects.
\textbf{2) Subject-Invariant Representation}: We extract subject-invariant semantic embeddings from disparate subjects' fMRI signals by utilizing a novel cycle reconstruction mechanism. These embeddings are then translated and aligned within a consistent CLIP embedding space, facilitating a standardized interpretation across varying neural responses.
\textbf{3) Efficient Adaptation Strategy}: To mitigate the data scarcity issue for new subjects, we introduce a novel finetuning method, reset-tuning. Because transferable knowledge from cross-subject pretraining is held in the deep layers, while the shallow layers are responsible for projecting diverse subjects' fMRI signals into subject-invariant embeddings. Reset-tuning reset the shallow layers but reuse the deep layers. 

Furthermore, MindBridge incorporates additional enhancements to improve semantic accuracy and expand its application. 
Utilizing a multi-modal versatile diffusion (VD) model \cite{xu2023versatile}, we can incorporate not only image stimuli but also the text caption as training data, then predict corresponding image and text embeddings to reconstruct more semantically faithful images.
Moreover, MindBridge opens new possibility to synthesize new brain signals using data from other subjects while preserving same semantic meaning by cyclic fMRI reconstruction. Therefore, MindBridge not only bridges the gap among different brains but also potentially augments the volume of available data.

To verify our approach, we conducted experiments on the publicly available NSD dataset \cite{allen2022massive}. Notably, the absence of common images across different subjects in the training set poses an additional challenge to cross-subject brain decoding. Surprisingly, MindBridge, employing only one model, achieves performance comparable to subject-specific methods, which require multiple models. Additionally, experiments on new subject adaptation validate that our method surpasses methods trained from scratch, showcasing the benefits of transferable knowledge from cross-subject pretraining and our proposed reset-tuning.

In summary, our contributions are as follows:

\begin{itemize}
    \item To the best of our knowledge, we are the first to effectively addresses the challenge of cross-subject brain decoding. We design a novel framework, MindBridge, equipped with an adaptive signal aggregation function and novel cyclic fMRI reconstruction mechanism for subject-invariant representations learning.
    \item We introduce a novel ``reset-tuning'' strategy, which efficiently adapts the MindBridge model to new subjects, and effectively overcoming the limitations posed by data scarcity for new subjects.
    \item MindBridge enables new capability for the synthesis of brain signals, leveraging data across various subjects while maintaining consistent semantic interpretation.
    \item Extensive experiments demonstrate MindBridge's efficacy and adaptability, showcasing its potential to significantly advance the field of brain decoding.
    
\end{itemize}
\vspace{-2mm}
\section{Related Work}
\label{sec:related}

\subsection{Brain Decoding}

The evolution of brain decoding has been marked by the integration of advanced modeling approaches. Earlier work \cite{horikawa2017generic} applies sparse linear regression on fMRI data to predict features from early convolutional layers of pretrained CNNs.
With the introduction of generative adversarial networks (GANs) \cite{goodfellow2020generative}, there has been a shift towards visual decoding techniques that map brain signals to the latent spaces of GANs, facilitating the reconstruction of handwritten digits \cite{schoenmakers2013linear}, human faces \cite{vanrullen2019reconstructing}, and natural scenes \cite{gu2022decoding, ozcelik2022reconstruction, seeliger2018generative}.

The advent of high-resolution image synthesis with Latent Diffusion Models \cite{rombach2022high} and multi-modal contrastive models like CLIP \cite{radford2021learning}, along with extensive fMRI datasets \cite{allen2022massive}, has propelled researchers to map fMRI signals into the CLIP embedding space. This mapping guides latent diffusion models for image reconstruction \cite{takagi2022high}, with efforts focusing on improved mapping through self-supervision \cite{beliy2019voxels}, masked modeling \cite{chen2023seeing}, and contrastive learning \cite{scotti2023reconstructing}. Additional explorations involve advanced diffusion models \cite{mai2023unibrain} and conditional control \cite{xia2023dream, lu2023minddiffuser}. 
Unlike almost all brain decoding research that requires training multiple models for different subjects, MindBridge stands out by aiming to achieve cross-subject brain decoding with a single model.

\subsection{Diffusion Models}

Diffusion models (DMs) \cite{ho2020denoising, song2020denoising, dhariwal2021diffusion,XingyiCVPR23,ma2023deepcache,wang2024patch,du2023stable} have recently emerged as a focal point in deep generative model research, known for their ability to generate high-quality images.
DMs utilize iterative denoising to recover a sampled variable from Gaussian noise and transform it into a sample conforming to the learned data distribution.
With large-scale image-text pair datasets \cite{schuhmann2022laion}, DMs have demonstrated superior performance in the task of text-to-image generation \cite{saharia2022photorealistic, ramesh2022hierarchical,XingyiICCV23,fang2023structural,pangu} and achieves unprecedented image quality.
Building on this, latent diffusion models (LDMs), also known as Stable Diffusion (SD), have furthered DMs by reducing computational demands through denoising in a latent space produced by autoencoders. An advanced form of LDMs, the Versatile Diffusion (VD) model \cite{xu2023versatile}, demonstrates the capability to produce high-quality images, guided by both image and text inputs. Consequently, we have adopted the VD model for its dual-input capacity, leveraging its enhanced image generation potential.


\begin{figure*}[th]
    \centering
    \includegraphics[width=1\linewidth]{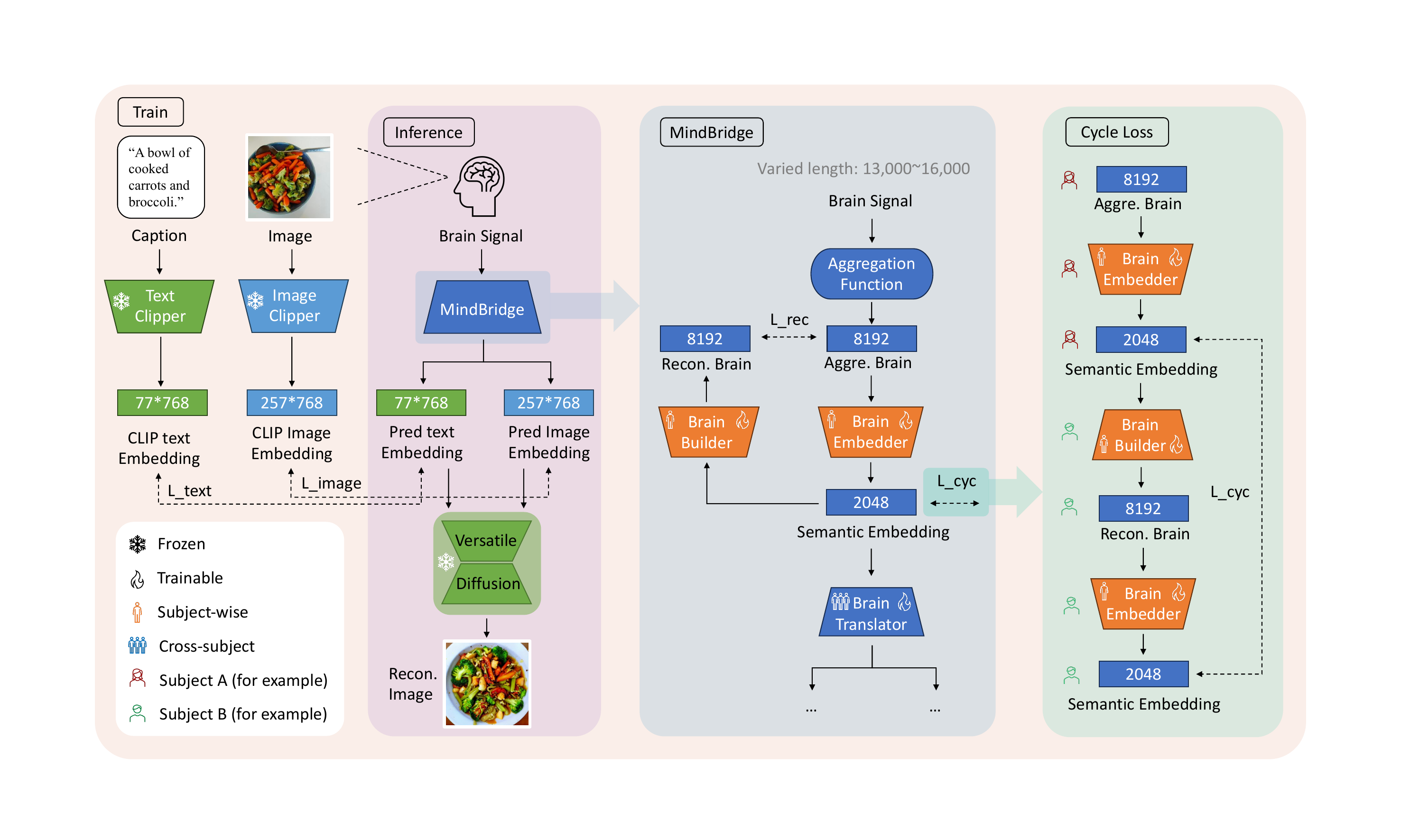}
    \caption{\textbf{Overview of MindBridge.} 
    MindBridge is a cross-subject brain decoding framework capable of handling fMRI signals from different subjects. Initially, an aggregation function unifies the size of fMRI signals. Subsequently, subject-wise brain embedders and brain builders are trained to obtain subject-invariant semantic embeddings. The Brain Translator then generates text and image embeddings, which are utilized to reconstruct images through versatile diffusion model.
    The dimension of data is denoted within the box. }
    \label{fig:enter-label}
\end{figure*}

\section{MindBridge}

\subsection{Data Elaboration}
\label{sec:data_elaboration}
To better understand the task at hand, we illustrate the data we used in ahead.
We have chosen the widely-used Natural Scenes Dataset (NSD) \cite{allen2022massive} for our brain decoding research. This dataset consists of high-resolution 7-Tesla fMRI scans collected from 8 healthy adult subjects, who were instructed to view thousands of natural images from MS-COCO dataset  \cite{lin2014microsoft}. 
Following common practices \cite{takagi2022high, scotti2023reconstructing, gu2022decoding, mai2023unibrain, ozcelik2023brain, xia2023dream}, our research mainly use data from 4 subjects (subj01, 02, 05, 07), who completed all the scan sessions.
Notably, only a subset of data, 982 images, were \textit{\textbf{commonly viewed}} by all four subjects. Those data were used as the \textit{\textbf{test set}}. While the remaining data, each of 8,859 \textbf{\textit{distinct}} images viewed by each subject were used as the \textit{\textbf{training set}}.
Following prior work \cite{scotti2023reconstructing}, we use preprocessed fMRI voxels from ``NSDGeneral'' regions of interest (ROI). 
Due to the inherent variablity in brain size and structure, the fMRI signals within the ROI exhibit different sizes (about 13,000 to 16,000 voxels per subject), which is the first challenge we need to tackle in cross-subject brain decoding.
The original acquired fMRI data is 4D (3D+t), which is firstly averaged among time dimension, and then is flattened from 3D to 1D. ROIs serves as masks on the 1D vector. So the dimensionality of input fMRI voxels is 1D.

\subsection{Cross-Subject Brain Decoding}
Current brain decoding pipeline can be summarized in two steps: mapping fMRI voxels to CLIP embeddings and then using these embeddings to guide diffusion models in generating reconstructed images. 
While previous brain decoding methods all fall in a per-subject-per-model fashion.
Here we argue the key insight for achieving cross-subject brain decoding lies in establishing a shared common representation space that is subject-invariant.
However, there are two main barriers to realizing this objective. The first is the variation in the size of fMRI signals among different subjects, as explained in Sec. \ref{sec:data_elaboration}. The second challenge is how to effectively model subject-invariant representation learning.

To address these challenges, we propose MindBridge, a novel framework for cross-subject brain decoding. Formally, we denote the 1D fMRI voxels from subject $s$ as $V_s \in \mathbb{R}^{F}$, where $F$ represents fMRI voxels' size.
The corresponding image stimulus $I$ and image caption $T$ can be extracted by a pretrained CLIP model as image embedding $e_I$ and text embedding $e_T$.

\textbf{Pipeline.} MindBridge first adaptively unifies the fMRI voxels $V_s$ to a unified size $v_s=f(V_s)$ using a biologically-inspired aggregation function $f$. Unlike previous methods that directly learn the projection between fMRI voxels and corresponding CLIP embeddings, MindBridge projects different subjects' aggregated fMRI voxels $v_s$ to an intermediate semantic embedding $e_s=\mathcal{E}_s(v_s)$ using a subject-wise brain embedder $\mathcal{E}_s$. 
To ensure that semantic embeddings from different subjects reside in a common shared space, we propose a novel cyclic fMRI reconstruction mechanism. 
This mechanism relys on an additional subject-wise brain builder $\mathcal{B}_s$ to reconstruct the unified fMRI voxels $\hat{v}_s=\mathcal{B}_s(e_s)$.
Once the semantic embeddings are obtained, a brain translator $\mathcal{T}$ translates them into two embeddings, $(\hat{e}_I, \hat{e}_T)=\mathcal{T}(e_s)$, 
representing the predicted CLIP image and text embeddings.
The brain embedder, brain builder and brain translator are all MLP-like networks. 

\textbf{Diffusion Model.}
Due to the limited volume of brain data, a generative model trained on a large-scale dataset is necessary to aid the image reconstruction process. Previous methods \cite{chen2023seeing, takagi2022high, scotti2023reconstructing} have demonstrated the superiority of using diffusion models as an interface for image generation. 
In this work, we have chosen to employ the versatile diffusion (VD) \cite{xu2023versatile} model, a multimodal latent diffusion model that is guided by image and text CLIP embeddings and achieves state-of-the-art performance in image generation. Its exceptional capabilities give us an opportunity to utilize both visual and semantic information, as represented by CLIP image and text embeddings predicted by MindBridge, to reconstruct images at inference time.

\textbf{Adaptive fMRI Aggregation.}
Modern neural-science research \cite{olshausen1996emergence, rao1999predictive, vinje2000sparse} reveals that visual stimuli are encoded sparsely in the primary visual cortex, activating only a few neurons for most natural images. 
Also, neurons require a certain level of threshold, to become active and fire an action potential \cite{hodgkin1952quantitative, kandel2000principles}. The activation functions of artificial neural networks such as Sigmoid or ReLU also echo this fundamental principle in neurophysiology \cite{goodfellow2016deep, rumelhart1986learning}.
Inspired by these findings, we posit that brain signals can be aggregated sparsely, with higher values tending to be more valuable. 
Consequently, we propose employing ``Adaptive Max Pooling" \footnote{PyTorch implementation: \url{https://pytorch.org/docs/stable/generated/torch.nn.AdaptiveMaxPool1d.html}} as the aggregation function. This function unifies the size of input fMRI signals by dynamically adjusting its pooling size to produce a fixed output size.


\textbf{Learning Objectives.}
MindBridge learns CLIP image and text embeddings through two types of losses.
One is the SoftCLIP loss, introduced in \cite{scotti2023reconstructing}, which has proven effective in aligning the fMRI modality with the embedding space of the pretrained CLIP model. This loss facilitates contrastive learning by maximizing the similarity of positive pairs while minimizing the similarity of negative pairs. Positive pairs are defined as the soft labels produced by the dot product of embeddings within a batch, and the loss considers both CLIP-CLIP and Brain-CLIP scenarios.

\vspace{-2mm}
\begin{gather}
    \mathcal{L}_{SoftCLIP}(p,t) = - \textstyle\sum_{i=1}^{N} \textstyle\sum_{j=1}^{N} \notag \\ 
    \left[
        \displaystyle\frac{\exp\left(t_i \cdot t_j / \tau\right)}
        {\sum_{m=1}^{N} \exp\left(\frac{t_i \cdot t_m}{\tau}\right)}
        \cdot \log 
        \left( 
            \displaystyle\frac{\exp\left(p_i \cdot t_j / \tau\right)}
            {\sum_{m=1}^{N} \exp\left(\frac{p_i \cdot t_m}{\tau}\right)}
        \right)
    \right]
\end{gather}
\vspace{2mm}

Where $p$, $t$ are the predicted CLIP embedding and target CLIP embedding in a batch of size $N$, respectively. $\tau$ is a temperature hyperparameter.

During our exploratory experiments, however, we observed that reconstructed images still exhibited some artifacts when using only the SoftCLIP loss. We hypothesize that this may be due to the SoftCLIP loss's inability to guarantee the authenticity of the learned CLIP embeddings. Therefore, we introduced the second loss, the MSE loss, to ensure a more accurate prediction of CLIP embeddings.

\vspace{-2mm}
\begin{equation}
\mathcal{L}_{MSE}(p,t) = \frac{1}{N}\textstyle\sum_{i=1}^{N}(p_i - t_i)^2
\end{equation}

Incorporating these two losses ensures a more natural image reconstruction. The complete set of losses for predicting image and text CLIP embeddings includes:

\vspace{-5mm}
\begin{align}
\mathcal{L}_{image} &= \mathcal{L}_{SoftCLIP}(\hat{e}_I,e_I) + \mathcal{L}_{MSE}(\hat{e}_I,e_I) \\
\mathcal{L}_{text} &= \mathcal{L}_{SoftCLIP}(\hat{e}_T,e_T) + \mathcal{L}_{MSE}(\hat{e}_T,e_T)
\end{align}

Where $e_I$ and $e_T$ are CLIP image and text embeddings of image stimuli $I$ and captions $T$.

\textbf{Cyclic fMRI Reconstruction.}
To facilitate subject-invariant representation learning, the simplest way is to directly minimize the distance between the semantic embeddings $e_s$ from two subjects when they are viewing the same images.
Nevertheless, as described in Sec. \ref{sec:data_elaboration}, there is no common-viewd image across different subjects in the training set. 
Therefore, we turn to design a mechanism, wishing to synthesize the fMRI signals even the subject does not actually see the image stimulus. In this way, we can mimic the scenario that two subjects are viewing the same images.





To realize that, we first introduce a brain builder $\mathcal{B}_s$ to reconstruct fMRI signal $\hat{v}_s=\mathcal{B}_s(\mathcal{E}_s(v_s))$ in a AutoEncoder \cite{bank2023autoencoders} manner. The reconstruction loss is:

\vspace{-2mm}
\begin{equation}
\mathcal{L}_{rec} = \frac{1}{N}\textstyle\sum_{i=1}^{N}(\hat{v}_s - v_s)^2
\end{equation}

We then randomly select two subjects $a$ and $b$ from all training subjects in every training iteration. 
Through a cyclic fMRI reconstruction, we can transform subject $a$'s fMRI signal $v_a$ into subject $b$'s $v_b=\mathcal{B}_b(\mathcal{E}_a(v_a))$ like they are viewing the same image.
This cycle is tenable only when the involved semantic embeddings $e_a=\mathcal{E}_a(v_a), e_b=\mathcal{E}_b(v_b)$ are really subject-invariant, that is, the same when viewing the same image. So a cycle loss is employed to ensure consistency in this cycle:

\vspace{-2mm}
\begin{equation}
\mathcal{L}_{cyc} = \frac{1}{N}\textstyle\sum_{i=1}^{N}(e_b - e_a)^2
\end{equation}

MindBridge is trained end-to-end by incorporating all these losses to achieve cross-subject brain decoding.

\vspace{-4mm}
\begin{equation}
\mathcal{L}_{total} = \mathcal{L}_{image} + \mathcal{L}_{text} + \mathcal{L}_{rec} + \mathcal{L}_{cyc}
\end{equation}

\begin{figure*}[t]
  \centering
  \includegraphics[width=\linewidth]{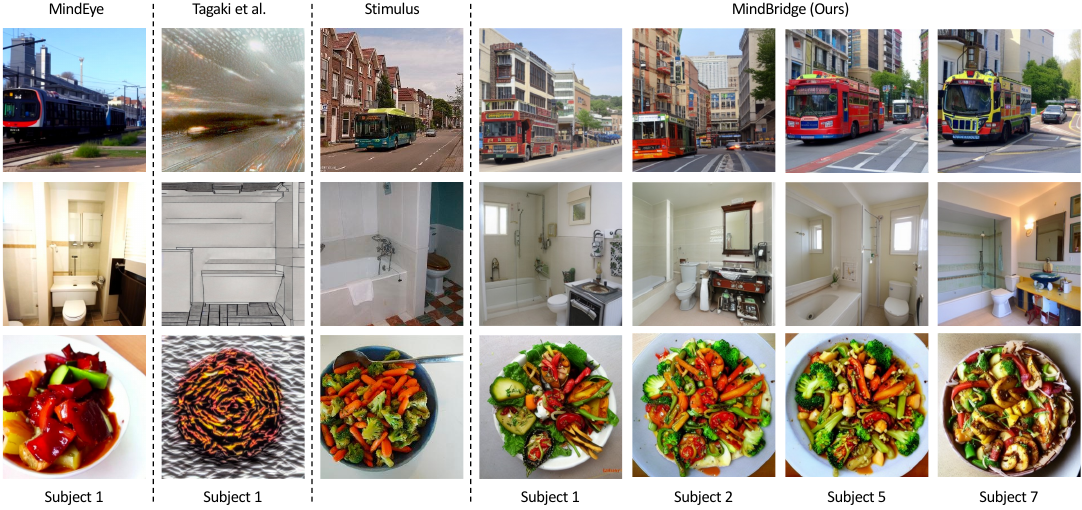}
  \caption{\textbf{Brain decoding results with only one model.} Unlike previous methods, which confine one model to a specific subject, our proposed cross-subject brain decoding framework, MindBridge, can reconstruct images from multiple subjects using just one model. }
  \label{fig:cross}
\end{figure*}

\begin{table*}[]
\centering
\begin{tabular}{l@{\hspace{0.3em}}cc@{\hspace{0.5em}}c@{\hspace{0.5em}}c@{\hspace{0.5em}}c@{\hspace{0.5em}}c@{\hspace{0.5em}}c@{\hspace{0.5em}}c@{\hspace{0.5em}}c@{\hspace{0.5em}}c}
    \toprule
    \multirow{2}[3]{*}{Method} & \multirow{2}[3]{*}{\# Models} & \multicolumn{4}{c}{Low-Level} & \multicolumn{4}{c}{High-Level} \\
    \cmidrule(lr){3-6} \cmidrule(lr){7-10} 
                                          &            & PixCorr $\uparrow$ & SSIM $\uparrow$ & Alex(2) $\uparrow$ & Alex(5) $\uparrow$ & Incep $\uparrow$ & CLIP $\uparrow$ & EffNet-B $\downarrow$ & SwAV $\downarrow$  \\
    \midrule
    Takagi et al. \cite{takagi2022high}    & 4          &     --                 &        --           &        83.0\%            &        83.0\%            &        76.0\%          &        77.0\%         &          --               &          --          \\
    Brain-Diffuser \cite{ozcelik2023brain} & 4          &         .254           &       .356          &         94.2\%           &             96.2\%       &        87.2\%          &        91.5\%         &          .775             &          .423          \\
    MindEye \cite{scotti2023reconstructing}& 4          &         .309           &        .323         &         94.7\%           &             97.8\%       &       93.8\%           &        94.1\%         &          .645             &         .367           \\
    MindBridge (Single)                   & 4          &         .148           &        .259         &         86.9\%           &         95.3\%           &       92.2\%           &        94.3\%         &          .713             &         .413           \\
    \midrule
    MindBridge (Ours)                     & \textbf{1} &         .151           &        .263         &         87.7\%           &        95.5\%            &        92.4\%          &      \textbf{94.7\%}  &          .712             &         .418           \\
    \bottomrule
\end{tabular}
\caption{\textbf{Quantitative comparison of brain decoding between MindBridge and other methods.} Our MindBridge is the first effective cross-subject brain decoding approach that only employs one model to reconstruct images from multiple subjects' fMRI signals. While other methods follows a per-subject-per-model fashion. All metrics are calculated as the average across 4 subjects.}
\label{tab:cross}
\end{table*}

\subsection{New-Subject Adaptation}
The scope of ``cross-subject'' is not limited to previously trained subjects. With MindBridge's ability to handle different subjects within a single model, adapting the model to a new subject is now feasible. This scenario is ubiquitous in real-world applications, such as diagnosing a new patient. However, in practice, acquiring brain signals for a new subject can be extremely costly and time-consuming. For example, the authors of NSD dataset \cite{allen2022massive} spent an entire year completing all fMRI scan sessions. To address the challenge of limited data for a new subject, we adopt the classic ``pretrain-then-finetune'' paradigm and propose two techniques: reset-tuning and pseudo data augmentation, to enhance the performance of new subject adaptation.

\textbf{Reset-Tuning.}
Contrary to traditional fine-tuning in computer vision, which often freezes shallow layers to leverage generic features \cite{yosinski2014transferable}, MindBridge adopts an inverse manner. The transferable knowledge within MindBridge resides in the deep layers, brain translator $\mathcal{T}$. While the shallow layers, brain embedder $\mathcal{E}_s$ and builder $\mathcal{B}_s$, are subject-specific due to human brain diversity. Hence, we propose reset-tuning strategy: training the brain embedder and builder from reset parameters while freezing the brain translator to retain the pretrained cross-subject knowledge.

\textbf{Pseudo Data Augmentation.}
A straightforward approach to mitigating the data scarcity problem is through data augmentation. However, suitable data augmentation method for brain signals is currently unavailable. Nevertheless, the cycle reconstruction mechanism of fMRI can serve as a form of pseudo data augmentation. 
During the adaptation process, fMRI signals from all previously trained subjects can be utilized to augment new subject's data: converted into the fMRI signals of the new subject through cycle reconstruction, regulated by $\mathcal{L}_{rec}$ and $\mathcal{L}_{cyc}$ too.

\section{Experiments and Analysis}

\textbf{Evaluation Metrics.} To quantitatively compare with other methods, we adopt eight image quality evaluation metrics following \cite{ozcelik2023brain}. PixCorr, SSIM \cite{wang2004image}, AlexNet(2), and AlexNet(5) \cite{krizhevsky2012imagenet} are used to evaluate low-level properties. Inception \cite{szegedy2016rethinking}, CLIP \cite{radford2021learning}, EffNet-B \cite{tan2019efficientnet}, and SwAV \cite{caron2020unsupervised} are considered for evaluating higher-level properties.


\begin{table*}[th]
\centering
\begin{tabular}{lcc@{\hspace{0.3em}}c@{\hspace{0.3em}}c@{\hspace{0.3em}}c@{\hspace{0.3em}}c@{\hspace{0.3em}}c@{\hspace{0.3em}}c@{\hspace{0.3em}}c@{\hspace{0.3em}}c}
    \toprule
    \multirow{2}[3]{*}{Method} & \multirow{2}[3]{*}{\# Data} & \multicolumn{4}{c}{Low-Level} & \multicolumn{4}{c}{High-Level} \\
    \cmidrule(lr){3-6} \cmidrule(lr){7-10} 
                           &         & PixCorr $\uparrow$ & SSIM $\uparrow$ & Alex(2) $\uparrow$ & Alex(5) $\uparrow$ & Incep $\uparrow$ & CLIP $\uparrow$ & EffNet-B $\downarrow$ & SwAV $\downarrow$  \\
    \midrule
    Vanilla   & 500     &    .079                &    .171             &       73.5\%             &        83.3\%            &        74.4\%          &       80.1\%          &        .894               &           .587         \\
    MindBridge (Ours)      & 500     &     \textbf{.112}               &     \textbf{.229}            &       \textbf{79.6\%}             &      \textbf{89.0\%}              &        \textbf{82.3\%}          &       \textbf{86.7\%}          &         \textbf{.840}              &           \textbf{.521}         \\
    \midrule
    Vanilla   & 1500    &       .107             &       .206          &       79.4\%             &       90.0\%             &        82.4\%          &       87.2\%          &         .844              &             .523       \\
    MindBridge (Ours)      & 1500    &       \textbf{.140}             &       \textbf{.250}          &              \textbf{84.6\%}      &        \textbf{92.6\%}            &        \textbf{85.8\%}          &       \textbf{91.0\%}          &         \textbf{.796}              &         \textbf{.485}           \\
    \midrule
    Vanilla   & 4000    &         .114           &         .232        &               81.4\%     &         92.2\%           &        85.3\%          &       89.8\%          &         .815              &         .491           \\
    MindBridge (Ours)      & 4000    &         \textbf{.156}           &       \textbf{.258}          &               \textbf{85.7\%}     &        \textbf{94.1\%}            &        \textbf{88.9\%}          &       \textbf{92.5\%}          &          \textbf{.765}             &          \textbf{.458}          \\
    \bottomrule
\end{tabular}
\caption{\textbf{Reseults of new subject adaptation in limited data scenario.} Here we report results from models that were trained on subsets of 500, 1500, and 4000 data points, selected from a total of 8859 training data points for subject 7. MindBridge (Ours) is fine-tuned using reset-tuning from the pretrained MindBridge model in subjects 1, 2, and 5. Compared to vanilla methods, which trains per-subject-per-model MindBridge from scratch, our MindBridge achieves superior brain decoding performance, benefiting from its use of pretrained cross-subject knowledge. For adaptation of other subjects, please refer to supplementary materials.}
\label{tab:new}
\end{table*}

\begin{table*}[h]
\centering
\begin{tabular}{lcccccccccc}
    \toprule
    \multirow{2}[3]{*}{Aggregation Function} & \multicolumn{4}{c}{Low-Level} & \multicolumn{4}{c}{High-Level} \\
    \cmidrule(lr){2-5} \cmidrule(lr){6-9} 
                         & PixCorr $\uparrow$ & SSIM $\uparrow$ & Alex(2) $\uparrow$ & Alex(5) $\uparrow$ & Incep $\uparrow$ & CLIP $\uparrow$ & EffNet-B $\downarrow$ & SwAV $\downarrow$  \\
    \midrule
    Interpolation        &      .151              &       .260          &        87.1\%            &         95.4\%           &        92.1\%          &        94.4\%         &           .712            &          .413          \\
    AdaAvgPool           &       .163             &        .274         &        87.4\%            &         95.7\%           &        92.8\%          &        94.5\%         &           .707            &          .405          \\
    AdaMaxPool (Ours)    &       \textbf{.165}            &         \textbf{.284}        &         \textbf{88.7\%}           &          \textbf{96.2\%}          &         \textbf{93.7\%}         &         \textbf{95.0\%}        &           \textbf{.697}            &           \textbf{.400}         \\
    \bottomrule
\end{tabular}
\caption{\textbf{Ablation of different aggregation functions.} Models are trained and evaluated on subject 1.}
\label{tab:aggre}
\end{table*}

\subsection{Cross-Subject Brain Decoding}
MindBridge can perform brain decoding for multiple subjects using one single model, whereas other methods require training separate models for each subject. 
To validate MindBridge's effectiveness, we compare its average image reconstruction performance across all four subjects with that of state-of-the-art methods: 
Takagi \textit{et al.} \cite{takagi2022high},
Brain-Diffuser \cite{ozcelik2023brain},
and MindEye \cite{scotti2023reconstructing}.
We also train a per-subject-per-model MindBridge for fair comparison, which is denoted as ``MindBridge (Single)''.
The quantitative and qualitative results for all methods are presented in Tab. \ref{tab:cross} and Fig. \ref{fig:cross} respectively.
Through subject-invariant representation learning, we achieve comparable performance against state-of-the-art methods while maintaining just one model, demonstrating our success on cross-subject brain decoding.

\begin{table*}[h]
\centering
\begin{tabular}{lc@{\hspace{0.9em}}c@{\hspace{0.9em}}c@{\hspace{0.9em}}c@{\hspace{0.9em}}c@{\hspace{0.9em}}c@{\hspace{0.9em}}c@{\hspace{0.9em}}c@{\hspace{0.9em}}c}
    \toprule
    \multirow{2}[3]{*}{Pretrain Loss} & \multicolumn{4}{c}{Low-Level} & \multicolumn{4}{c}{High-Level} \\
    \cmidrule(lr){2-5} \cmidrule(lr){6-9} 
                         & PixCorr $\uparrow$ & SSIM $\uparrow$ & Alex(2) $\uparrow$ & Alex(5) $\uparrow$ & Incep $\uparrow$ & CLIP $\uparrow$ & EffNet-B $\downarrow$ & SwAV $\downarrow$  \\
    \midrule
    SoftCLIP loss                   &       .085             &       \textbf{.336}          &        76.9\%            &         83.1\%           &        79.1\%          &        80.6\%         &             .877          &          .542          \\
    + MSE loss                      &       .158             &       .272          &        88.3\%            &         95.7\%           &        92.2\%          &        94.5\%         &             .720          &          .418          \\
    + Recon + Cycle Loss (Ours)     &       \textbf{.168}             &       .277          &        \textbf{88.7\%}            &         \textbf{96.1\%}           &        \textbf{92.7\%}          &        \textbf{94.9\%}         &             \textbf{.707}          &          \textbf{.410}          \\
    \bottomrule
\end{tabular}
\caption{\textbf{Ablation of different losses at pretraining stage.} Models are trained on subject 1,2 and 5, then evaluated on subject 1. }
\label{tab:loss}
\end{table*}

\begin{table*}[h]
\centering
\begin{tabular}{l@{\hspace{0.9em}}c@{\hspace{0.9em}}c@{\hspace{0.9em}}c@{\hspace{0.9em}}c@{\hspace{0.9em}}c@{\hspace{0.9em}}c@{\hspace{0.9em}}c@{\hspace{0.9em}}c@{\hspace{0.9em}}c}
    \toprule
    \multirow{2}[3]{*}{Finetune Strategy} & \multicolumn{4}{c}{Low-Level} & \multicolumn{4}{c}{High-Level} \\
    \cmidrule(lr){2-5} \cmidrule(lr){6-9} 
                                      & PixCorr $\uparrow$ & SSIM $\uparrow$ & Alex(2) $\uparrow$ & Alex(5) $\uparrow$ & Incep $\uparrow$ & CLIP $\uparrow$ & EffNet-B $\downarrow$ & SwAV $\downarrow$  \\
    \midrule
    Full-tuning + $\mathcal{L}_{image}$ + $\mathcal{L}_{text}$        &       .110             &        \textbf{.232}         &          79.5\%          &         88.1\%           &        79.6\%          &       86.4\%          &         .847              &         .526           \\
    Full-tuning + $\mathcal{L}_{total}$          &         .100           &        .220         &          78.6\%          &         88.0\%           &        79.8\%          &       86.0\%          &          .851             &                .529    \\
    Reset-tuning + $\mathcal{L}_{image}$ + $\mathcal{L}_{text}$       &        \textbf{.116}            &        .227         &          \textbf{79.7\%}          &         88.9\%           &        80.5\%          &        86.1\%         &         .851              &         .525           \\
    Reset-tuning + $\mathcal{L}_{total}$ (Ours)   &         .112           &        .229         &         79.6\%           &         \textbf{89.0\%}           &        \textbf{82.3\%}          &        \textbf{86.7\%}         &           \textbf{.840}            &              \textbf{.520}      \\
    \bottomrule
\end{tabular}
\caption{\textbf{Ablation of different finetuning strategies.} Models are trained on subject 1,2 and 5, then finetuned and evaluated on subject 7.}
\label{tab:finetune}
\end{table*}

\subsection{New Subject Adaptation}
MindBridge also possesses a capability to transfer its pretrained knowledge for adapting to new subjects, which is valuable in practical applications where collecting brain signals for new subjects is resource-intensive and time-consuming.
To simulate scenarios with limited data, we tested our method using subsets of the total 8859 training data – specifically, 500, 1500, and 4000 data points – for new subject adaptation. We selected three subjects for pretraining (source subjects) and one additional subject for adaptation (target subject).
We choose to compare our method with the ``vanilla" approach, which involves training MindBridge from ``scratch" on the same target data in a per-subject-per-model fashion.
In Fig. \ref{fig:teaser}, we present a qualitative comparison of our method with vanilla method. The vanilla method struggles to reconstruct reasonable images, which can be attributed to two main factors. 
Firstly, our pretrained brain translator serves as a robust prior backbone, transferring highly useful knowledge that significantly enhances our method's performance. 
Secondly, the full-parameter model used in the vanilla approach tends to overfit when data is limited. In contrast, our approach employs reset-tuning, updating only the parameters within the lightweight brain embedder and brain builder, effectively preventing overfitting.
The quantitative comparison shown in Tab. \ref{tab:new} demonstrates that our method not only significantly outperforms traditional approaches but also highlights the feasibility of reliable brain decoding with substantially less data. 
This advancement opens up exciting prospects for considerably reducing scan times in practical applications, paving the way for more cost-efficient and generalizable brain decoding strategies.

\begin{figure}
  \centering
  \includegraphics[width=\linewidth]{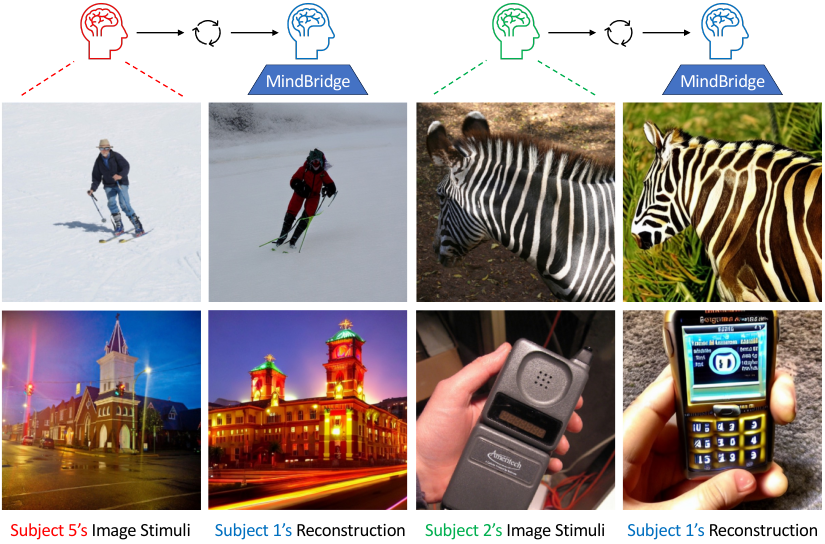}
  \caption{\textbf{Novel fMRI synthesis within MindBridge pretrained on subject 1, 2, 5}. The fMRI signals of subjects 5 and 2 are converted into subject 1’s fMRI signals through cycle reconstruction, then subject 1’s brain embedder are utilized for brain decoding.}
  \label{fig:novel}
\end{figure}

\subsection{Novel fMRI Synthesis}
Utilizing our cycle reconstruction mechanism, we have enabled a new task: novel fMRI synthesis. This process can transform one's fMRI signal into another's, while preserving the same semantic content as the original stimuli. By employing the pretrained MindBridge model on subject 1,2, and 5, we converted fMRI signals of subject 5 and 2 into those of subjects 1 using their respective brain embedders and brain builders. To validate the quality of these novel fMRI signals, we display the reconstructed images from the synthesized novel fMRI signals in Fig. \ref{fig:novel}. Notably, the stimuli corresponding to these novel fMRI signals have never been viewed by subject 1. Yet, they can still be reconstructed faithfully, demonstrating the effectiveness of our proposed cycle reconstruction mechanism in synthesize fMRI as well as facilitating subject-invariant representation learning.

\subsection{Ablation Study}

\textbf{Ablation on Aggregation Functions.}
A key component in cross-subject brain decoding is the aggregation function. The ability of this function to retain valuable information while unifying the dimensions of brain signals is crucial. The more effectively it preserves useful information, the more accurate the results will be during the brain decoding process.
We show ablation comparison with other functions in Tab. \ref{tab:aggre}.
Compared to adaptive average pooling and interpolation functions, our chosen function, adaptive max pooling, not only offers better biological interpretability but also achieves superior performance.

\textbf{Ablation on Pretraining Losses.}
We present the results of involving different losses at the pretraining stage in Tab. \ref{tab:loss}. When only the SoftCLIP loss is applied, the model struggles to fully learn the reasonable CLIP embeddings and can only achieve a resemblance to the target CLIP embeddings. The inclusion of MSE loss enhances the naturalness of the reconstructed images. Finally, the addition of both reconstruction loss and cycle loss improves the integrity of subject-invariant representation learning, thereby enhancing cross-subject brain decoding performance.

\textbf{Ablation on Finetuning Strategies.}
Once we have obtained the pretrained model, aside from our proposed reset-tuning strategy, we have several options for fine-tuning it to adapt to a new subject. Full-tuning involves fine-tuning the brain translator, while reset-tuning entails keeping the brain translator frozen. Both fine-tuning methods involve training a new brain embedder and brain builder, if applicable. We also conducted an ablation study of losses during fine-tuning to assess the benefits of using pseudo data augmentation, which corresponds to the application of losses related to cycle reconstruction.
Tab. \ref{tab:finetune} presents a quantitative comparison among these fine-tuning strategies. The results indicate that the strategy combining reset-tuning with pseudo data augmentation yields the most satisfactory results. This outcome suggests that reset-tuning, which only establishes a projection between brain signals and semantic embeddings, can already sufficiently adapt to a new subject. Moreover, the incorporation of our novel pseudo data augmentation can further imporove performance.

\section{Discussion}
Currently, due to the limited availability of high-quality fMRI data, one limitation of our paper is the evaluation is only restricted to a small dataset NSD. As a cross-subject framework, the generalizability could be further validated on a more diverse and large-scale dataset in the future. 
While considering the high cost of acquiring fMRI data, our method also offers a potential solution to reduce scan time. 
Another limitation is that the fMRI signals are serialized as 1D vectors, which may ruin the original spatial relationship.

Although brain decoding holds promise for assisting visual impaired people, ethical concerns arise regarding misuse for malicious or immoral purposes. Thus, a consented data privacy protocol and a responsible research code of conduct must be established with broader considerations.

\section{Conclusion}

In this paper, we introduced ``MindBridge'', a novel cross-subject brain decoding framework that successfully challenges the conventional per-subject-per-model paradigm in brain decoding. By innovatively addressing the critical issues of size variability, diverse neural responses, and data scarcity for new subjects, MindBridge demonstrates significant advancements in cross-subject brain decoding. Our approach, characterized by adaptive signal aggregation, cyclic fMRI reconstruction for subject-invariant representation, and reset-tuning for new subject adaptation, has proven effective in our experiments with the NSD dataset. These achievements not only enhance the decoding accuracy across multiple subjects but also open new avenues for fMRI synthesis and practical applications in neuroscience.

\section{Acknowledgement}
This project is supported by the Ministry of Education Singapore, under its Academic Research Fund Tier 2 (Award Number: MOE-T2EP20122-0006).

{
    \small
    \bibliographystyle{ieeenat_fullname}
    \bibliography{main}
}

\newpage
\setcounter{section}{0}

\renewcommand\thesection{\Alph{section}}
\clearpage
\setcounter{page}{1}
\maketitlesupplementary

\section{More Dataset Information} 
We used the Natural Scenes Dataset (NSD)\cite{allen2022massive} for all experiments. 
This dataset consists of high-resolution 7-Tesla fMRI scans collected from 8 healthy adult subjects. Every subject was instructed to view thousands of natural images from MS-COCO dataset\cite{lin2014microsoft} over the course of 30–40 scan sessions.
Each participant was exposed to 9,000-10,000 distinct images, each displayed for three seconds and repeated three times, resulting in a total of 22,000-30,000 fMRI response trials per participant. The fMRI responses were derived from GLMSingle, with outputs being session-wise z-scored single-trial betas \cite{prince2022improving}.
Following common practices \cite{takagi2022high, scotti2023reconstructing, gu2022decoding, mai2023unibrain, ozcelik2023brain, xia2023dream}, our research mainly use data from 4 subjects (subj01, subj02, subj05, subj07), who completed all the scan sessions, as our experimental data. 
Following prior work \cite{scotti2023reconstructing}, we use preprocessed fMRI voxels from ``NSDGeneral'' regions of interest (ROI). 
This ROI is defined by NSD authors as a subset of voxels in the posterior cortex that respond most strongly to the visual stimuli presented. 

Each subject was instructed to view up to three repeated trials per image. For the training set, we randomly choose one single-trial fMRI signal per image for training. 
For the test set, we average fMRI signals across the three same-image repetitions for the test set, similar to \cite{takagi2022high, scotti2023reconstructing}.

\section{Implementation Details} 
Our method's image reconstruction part relies on versatile diffusion model\cite{xu2023versatile}. 
In the diffusion process, we employ the UniPCMultistep scheduler\cite{zhao2023unipc} to execute 20 diffusion steps with a guidance scale of 5, setting the text-image ratio at 0.5 to integrate both visual and semantic information. For each test sample, we reconstruct 8 images and select the one with the highest CLIP similarity to the stimuli as the final result. 

We implemented the aggregation function as AdaptiveMaxPool1D in our PyTorch code, setting the output size to $8192$, which is the largest power-of-two integer smaller than all the fMRI signal sizes. In practice, we incorporated AdaptiveMaxPool1D into the data preprocessing pipeline of the dataloader to ensure uniform data dimensions within each batch.
We use the last hidden layer of CLIP ViT-L/14 for the CLIP image embedding and CLIP text embedding, results a shape of $257\times768$ and $77\times768$ respectively. 

The total loss is balanced by setting the weights of $\mathcal{L}_{image}$, $\mathcal{L}_{text}$, $\mathcal{L}_{rec}$, and $\mathcal{L}_{cyc}$ to $1, 1e4, 1$, and 1, respectively. The MindBridge models for cross-subject experiments are trained for 600 epochs, and those for new-subject adaptation are trained for 200 epochs. Across all experiments, the batch size is set to 50 per GPU.
Since $\mathcal{L}_{rec}$ and $\mathcal{L}_{cyc}$ are designed for subject-invariant representation learning, we do not employ them in single-subject model training. Therefore, we only use $\mathcal{L}_{image}$ and $\mathcal{L}_{text}$ for single-subject model training.

Data augmentation are applied to target images in all experiments to enhance model robustness. These include random cropping, resizing, and random adjustments of brightness, contrast, gamma, saturation, hue, sharpness, and gray scale.
In the pseudo data augmentation for new subject adaptation, we use the same number of data points from a previously trained subject as that of the new subject. And all previously trained subjects are involved in the training process.

\begin{figure}[t]
  \centering
  \includegraphics[width=\linewidth]{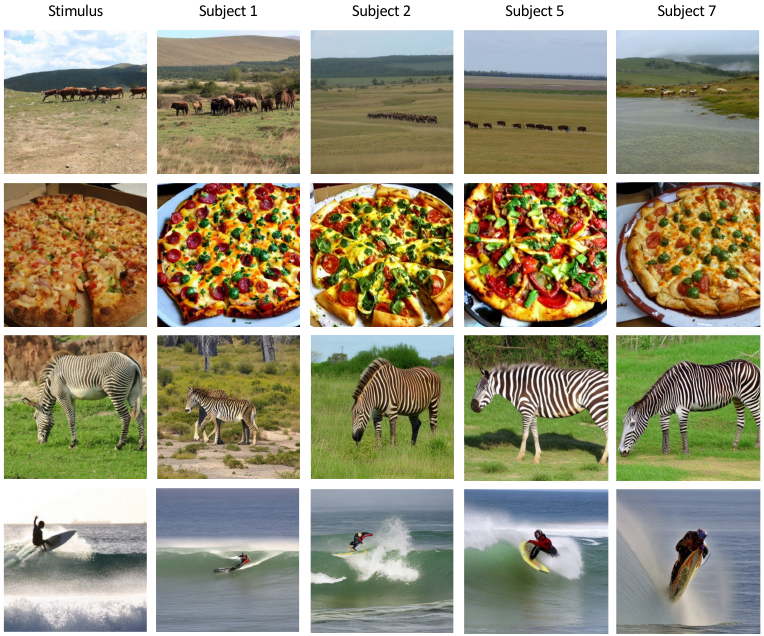}
  \caption{More cross-subject reconstructions of MindBridge on subject 1, 2, 5 and 7.}
  \label{fig:more}
\end{figure}

\section{More Reconstruction Results}

Here we show more visual results of cross-subject reconstruction from brain signals using one MindBridge model trained on subject 1, 2, 5 and 7 in Figure \ref{fig:more}.

\section{Architecture of MindBridge}
MindBridge comprises an aggregation function, a brain embedder, a brain builder, and a brain translator. The aggregation function is adaptive max pool. The brain embedder consists of a LoRA\cite{hu2021lora}-like adapter and one linear layer. The brain builder is basically the reverse of brain embedder. The brain translator is a 4-layer residual MLP equipped with two linear heads, and the hidden layer size is 2048.

The PyTorch-like architectural code for the brain embedder, brain builder, and brain translator is illustrated in Fig \ref{fig:alg}.

\begin{figure*}[h]
\begin{lstlisting}
class MindBridge(nn.Module):
    def __init__(self, in_dim=8196, out_dim_image=257*768, out_dim_text=77*768, h=2048, subj_list=[1,2,5,7])
        self.embedder = nn.ModuleDict(subj: nn.Sequential(
            Adapter(in_dim, 128),
            nn.Linear(in_dim, h),
            nn.LayerNorm(h), 
            nn.GELU(),
            nn.Dropout(0.5)
        ) for subj in subj_list})

        self.builder = nn.ModuleDict({subj: nn.Sequential(
            nn.Linear(h, in_dim),
            nn.LayerNorm(in_dim),
            nn.GELU(),
            Adapter(in_dim, 128)
        ) for subj in subj_list})

        self.mlp = nn.ModuleList([
            nn.Sequential(
                nn.Linear(h, h),
                nn.LayerNorm(h), 
                nn.GELU(),
                nn.Dropout(0.15)
        ) for _ in range(4)])

        self.head_image = nn.Linear(h, out_dim_image)
        self.head_text  = nn.Linear(h, out_dim_text)

\end{lstlisting}
\caption{PyTorch-like pseudo code of MindBridge architecture.}
\label{fig:alg}
\end{figure*}

\begin{table*}[h]
\centering
\begin{tabular}{lcc@{\hspace{0.3em}}c@{\hspace{0.3em}}c@{\hspace{0.3em}}c@{\hspace{0.3em}}c@{\hspace{0.3em}}c@{\hspace{0.3em}}c@{\hspace{0.3em}}c@{\hspace{0.3em}}c}
    \toprule
    \multirow{2}[3]{*}{Method} & \multirow{2}[3]{*}{\# Data} & \multicolumn{4}{c}{Low-Level} & \multicolumn{4}{c}{High-Level} \\
    \cmidrule(lr){3-6} \cmidrule(lr){7-10} 
                           &         & PixCorr $\uparrow$ & SSIM $\uparrow$ & Alex(2) $\uparrow$ & Alex(5) $\uparrow$ & Incep $\uparrow$ & CLIP $\uparrow$ & EffNet-B $\downarrow$ & SwAV $\downarrow$  \\
    \midrule
    MindBridge (Scratch)   & 500     &    .097                &    .199             &       76.9\%             &        87.3\%            &        75.7\%          &       82.8\%          &        .885               &           .560         \\
    MindBridge (Ours)      & 500     &     \textbf{.142}               &     \textbf{.263}            &       \textbf{84.4\%}             &      \textbf{91.4\%}              &        \textbf{82.3\%}          &       \textbf{89.4\%}          &         \textbf{.831}              &           \textbf{.500}         \\
    \midrule
    MindBridge (Scratch)   & 1500    &       .122             &       .235          &       81.6\%             &       91.5\%             &        82.8\%          &       87.7\%          &         .838              &             .506       \\
    MindBridge (Ours)      & 1500    &       \textbf{.161}             &       \textbf{.275}          &              \textbf{87.2\%}      &        \textbf{94.2\%}            &        \textbf{88.5\%}          &       \textbf{92.5\%}          &         \textbf{.784}              &         \textbf{.460}           \\
    \midrule
    MindBridge (Scratch)   & 4000    &         .138           &         .266        &               85.8\%     &         94.3\%           &        87.8\%          &       91.5\%          &         .800              &         .465           \\
    MindBridge (Ours)      & 4000    &         \textbf{.157}           &       \textbf{.275}          &               \textbf{88.1\%}     &        \textbf{95.5\%}            &        \textbf{90.0\%}          &       \textbf{93.9\%}          &          \textbf{.747}             &          \textbf{.436}          \\
    \bottomrule
\end{tabular}
\caption{\textbf{Results of new subject adaptation in limited data scenario.} MindBridge(Ours) is finetuned on subject 1 from model pretrained on subject 2, 5 and 7.}
\label{tab:new1}
\end{table*}

\begin{table*}[h]
\centering
\begin{tabular}{lcc@{\hspace{0.3em}}c@{\hspace{0.3em}}c@{\hspace{0.3em}}c@{\hspace{0.3em}}c@{\hspace{0.3em}}c@{\hspace{0.3em}}c@{\hspace{0.3em}}c@{\hspace{0.3em}}c}
    \toprule
    \multirow{2}[3]{*}{Method} & \multirow{2}[3]{*}{\# Data} & \multicolumn{4}{c}{Low-Level} & \multicolumn{4}{c}{High-Level} \\
    \cmidrule(lr){3-6} \cmidrule(lr){7-10} 
                           &         & PixCorr $\uparrow$ & SSIM $\uparrow$ & Alex(2) $\uparrow$ & Alex(5) $\uparrow$ & Incep $\uparrow$ & CLIP $\uparrow$ & EffNet-B $\downarrow$ & SwAV $\downarrow$  \\
    \midrule
    MindBridge (Scratch)   & 500     &    .081                &    .200             &       76.0\%             &        87.8\%            &        77.3\%          &       81.6\%          &        .884               &           .546         \\
    MindBridge (Ours)      & 500     &     \textbf{.122}               &     \textbf{.265}            &       \textbf{83.2\%}             &      \textbf{91.0\%}              &        \textbf{83.0\%}          &       \textbf{87.2\%}          &         \textbf{.833}              &           \textbf{.501}         \\
    \midrule
    MindBridge (Scratch)   & 1500    &       .107             &       .234          &       82.0\%             &       93.2\%             &        82.7\%          &       87.1\%          &         .835              &             .497       \\
    MindBridge (Ours)      & 1500    &       \textbf{.143}             &       \textbf{.273}          &              \textbf{86.6\%}      &        \textbf{94.2\%}            &        \textbf{88.2\%}          &       \textbf{91.7\%}          &         \textbf{.780}              &         \textbf{.459}           \\
    \midrule
    MindBridge (Scratch)   & 4000    &         .128           &         .257        &               85.5\%     &         94.2\%           &        87.1\%          &       90.1\%          &         .801              &         .469           \\
    MindBridge (Ours)      & 4000    &         \textbf{.150}           &       \textbf{.272}          &               \textbf{88.4\%}     &        \textbf{95.5\%}            &        \textbf{89.8\%}          &       \textbf{93.0\%}          &          \textbf{.745}             &          \textbf{.436}          \\
    \bottomrule
\end{tabular}
\caption{\textbf{Results of new subject adaptation in limited data scenario.} MindBridge(Ours) is finetuned on subject 2 from model pretrained on subject 1, 5 and 7.}
\label{tab:new2}
\end{table*}

\begin{table*}[h]
\centering
\begin{tabular}{lcc@{\hspace{0.3em}}c@{\hspace{0.3em}}c@{\hspace{0.3em}}c@{\hspace{0.3em}}c@{\hspace{0.3em}}c@{\hspace{0.3em}}c@{\hspace{0.3em}}c@{\hspace{0.3em}}c}
    \toprule
    \multirow{2}[3]{*}{Method} & \multirow{2}[3]{*}{\# Data} & \multicolumn{4}{c}{Low-Level} & \multicolumn{4}{c}{High-Level} \\
    \cmidrule(lr){3-6} \cmidrule(lr){7-10} 
                           &         & PixCorr $\uparrow$ & SSIM $\uparrow$ & Alex(2) $\uparrow$ & Alex(5) $\uparrow$ & Incep $\uparrow$ & CLIP $\uparrow$ & EffNet-B $\downarrow$ & SwAV $\downarrow$  \\
    \midrule
    MindBridge (Scratch)   & 500     &    .078                &    .194             &       75.0\%             &        88.7\%            &        81.3\%          &       86.7\%          &        .846               &           .517         \\
    MindBridge (Ours)      & 500     &     \textbf{.124}               &     \textbf{.255}            &       \textbf{83.1\%}             &      \textbf{91.9\%}              &        \textbf{86.2\%}          &       \textbf{91.2\%}          &         \textbf{.794}              &           \textbf{.477}         \\
    \midrule
    MindBridge (Scratch)   & 1500    &       .107             &       .228          &       80.9\%             &       92.4\%             &        85.8\%          &       90.8\%          &         .804              &             .481       \\
    MindBridge (Ours)      & 1500    &       \textbf{.135}             &       \textbf{.262}          &              \textbf{86.1\%}      &        \textbf{93.8\%}            &        \textbf{89.8\%}          &       \textbf{93.5\%}          &         \textbf{.746}              &         \textbf{.440}           \\
    \midrule
    MindBridge (Scratch)   & 4000    &         .123           &         .239        &               84.2\%     &         94.6\%           &        90.1\%          &       93.6\%          &         .760              &         .442           \\
    MindBridge (Ours)      & 4000    &         \textbf{.150}           &       \textbf{.267}          &               \textbf{87.9\%}     &        \textbf{95.6\%}            &        \textbf{91.9\%}          &       \textbf{95.0\%}          &          \textbf{.712}             &          \textbf{.418}          \\
    \bottomrule
\end{tabular}
\caption{\textbf{Results of new subject adaptation in limited data scenario.} MindBridge(Ours) is finetuned on subject 5 from model pretrained on subject 1, 2 and 5.}
\label{tab:new5}
\end{table*}

\begin{figure*}[th]
  \centering
  \includegraphics[width=\linewidth]{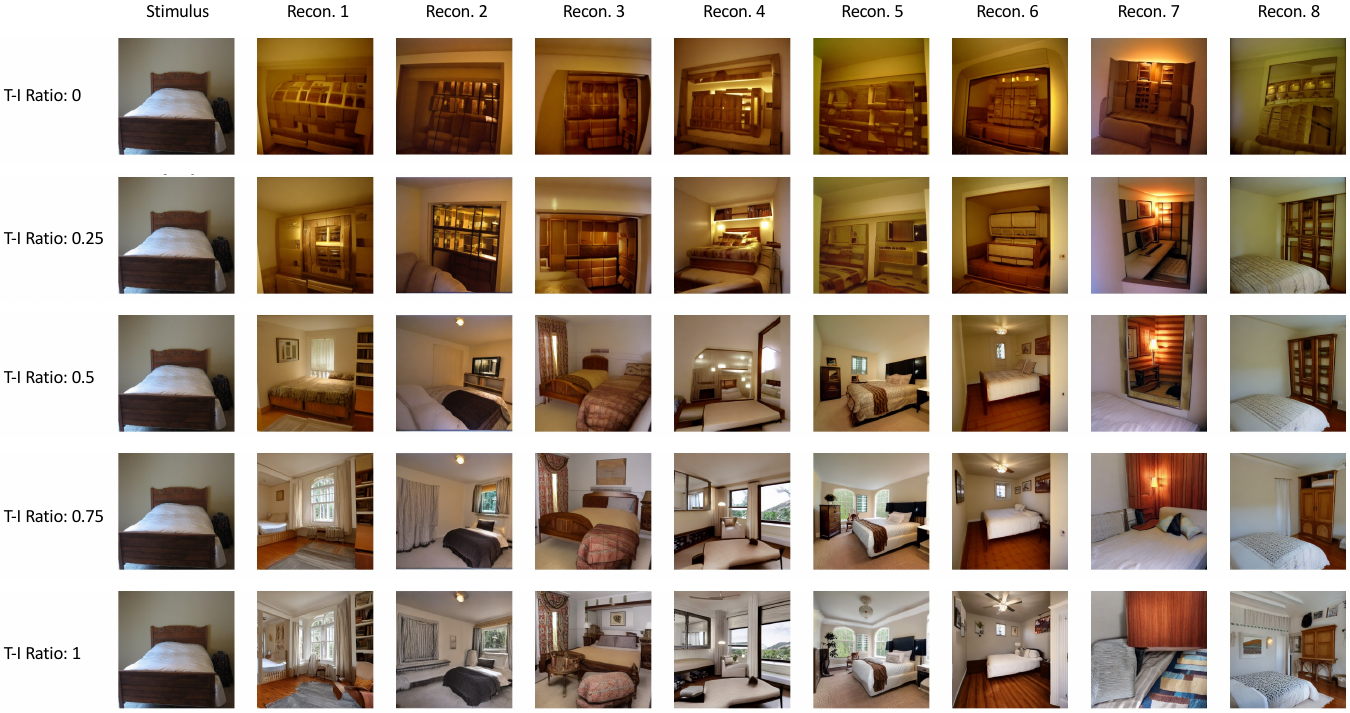}
  \caption{Influence of different text-image ratio. Model is trained and tested on subject 1. We reconstruct 8 images using 8 random seeds for illustration.}
  \label{fig:ratio}
\end{figure*}

\section{More Results for New Subject Adaptation}
Here, we present additional results of new subject adaptation for subjects 1, 2, and 5 in Table \ref{tab:new1}, Table \ref{tab:new2}, and Table \ref{tab:new5}, respectively.
The results show that our method consistently yields superior performance compared to the vanilla method.

\section{Effect of Text-Image Ratio}
Because versatile diffusion\cite{xu2023versatile} model can accept both image and text inputs, we demonstrate how varying the image-text ratio can influence the trade-off between image consistency and diversity in Figure \ref{fig:ratio}. A greater proportion of text input ensures more semantic correctness while introducing more diversity. To balance semantic accuracy and visual consistency, we have chosen to set the image-text ratio to $0.5$ in this work.

\section{Ethic and Social Impact}
As brain decoding technology advances, it brings critical ethical considerations to the forefront. While it promises enhanced communication for those with speech or motor impairments, its potential for involuntary mind reading necessitates stringent ethical frameworks. Key elements include informed consent, robust data privacy, and a thorough consideration of societal implications. It is crucial for the scientific community and society to address these ethical challenges to prevent misuse and ensure the responsible development of brain decoding technology.

\end{document}